\documentclass[10pt,twocolumn,letterpaper]{article}

\usepackage{cvpr}
\usepackage{times}
\usepackage{epsfig}
\usepackage{graphicx}
\usepackage{amsmath,amssymb,amsfonts}
\usepackage{algorithm}
\usepackage{algpseudocode}
\usepackage{mathtools}
\usepackage{booktabs}
\usepackage{caption}
\usepackage{subcaption}
\usepackage{pifont}
\usepackage{amssymb}
\usepackage{balance}

\usepackage[pagebackref=true,breaklinks=true,letterpaper=true,colorlinks,bookmarks=false]{hyperref}

\newcommand{\myparagraph}[1]{\vspace{4pt} \noindent \textbf{#1.}}
\newcommand{\argmin}[1]{\underset{#1}{\operatorname{arg}\,\operatorname{min}}\;}

\newcommand{\argrelmin}[1]{\underset{#1}{\operatorname{arg}\,\operatorname{rel}\,\operatorname{min}}\;}

\cvprfinalcopy 

\def\cvprPaperID{34} 

\ifcvprfinal\fi
\begin{document}

\title{Seeing Red: PPG Biometrics Using Smartphone Cameras}

\author{Giulio Lovisotto, Henry Turner, Simon Eberz and Ivan Martinovic\\
University of Oxford, UK\\
{\tt\small name.surname@cs.ox.ac.uk}
}

\maketitle

\begin{abstract}
In this paper, we propose a system that enables photoplethysmogram (PPG)-based authentication by using a smartphone camera.
PPG signals are obtained by recording a video from the camera as users are resting their finger on top of the camera lens. 
The signals can be extracted based on subtle changes in the video that are due to changes in the light reflection properties of the skin as the blood flows through the finger.
We collect a dataset of PPG measurements from a set of 15 users over the course of 6-11 sessions per user using an iPhone X for the measurements. 
We design an authentication pipeline that leverages the uniqueness of each individual's cardiovascular system, identifying a set of distinctive features from each heartbeat.
We conduct a set of experiments to evaluate the recognition performance of the PPG biometric trait, including cross-session scenarios which have been disregarded in previous work.
We found that when aggregating sufficient samples for the decision we achieve an EER as low as 8\%, but that the performance greatly decreases in the cross-session scenario, with an average EER of 20\%.
\end{abstract}

%

\section{Introduction}

Biometric authentication is a popular approach to effortless user authentication, especially on mobile devices such as smartphones or tablets. While both fingerprint scanning and face recognition offer quick and accurate authentication, their modalities are also easy for attackers to observe. At the same time, reliably detecting forged biometric samples (such as latex fingers) remains an arms race with attackers crafting increasingly sophisticated forgeries.
\begin{figure}[t]
	\centering
	\includegraphics[width=.95\columnwidth]{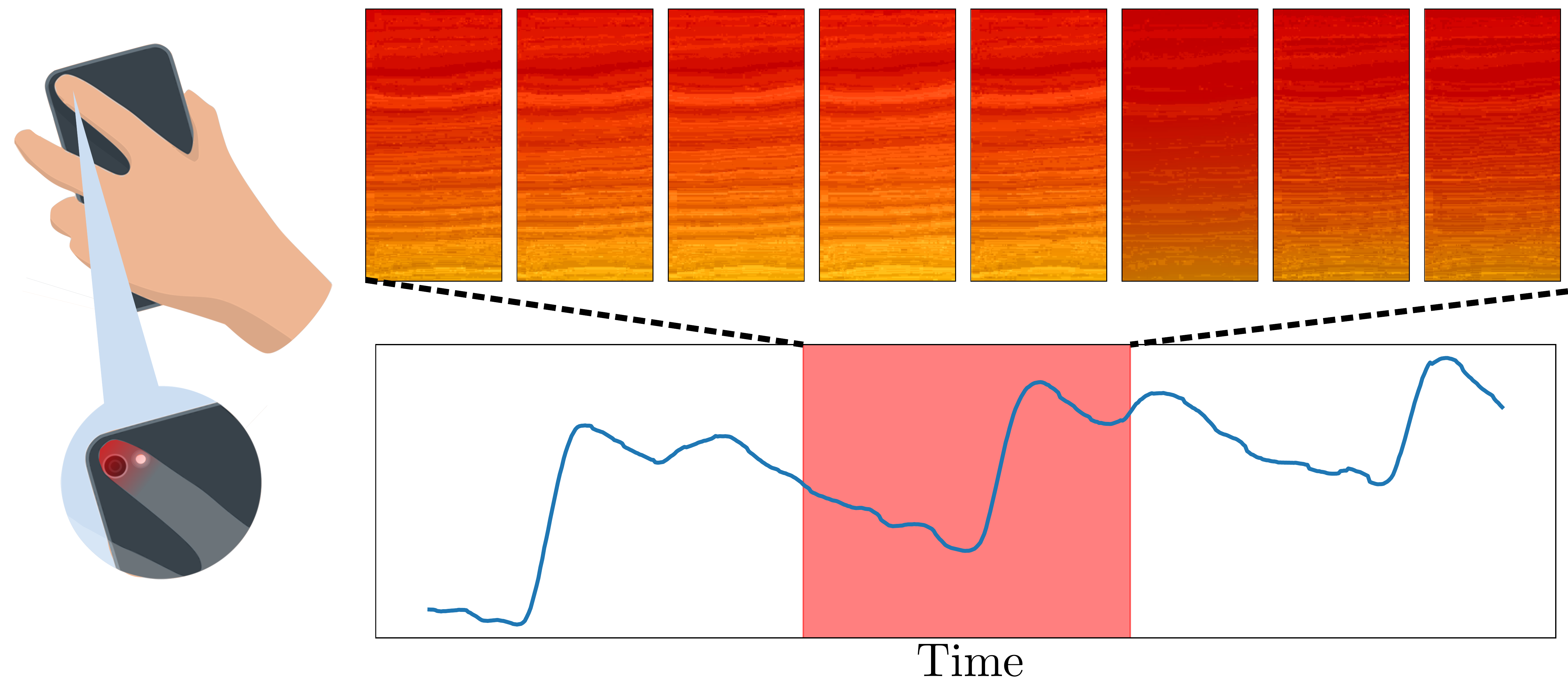}
	\caption{System Overview. Users place their finger on the smartphone camera. The blood flow to the fingertip changes the light reflection properties of the skin creating subtle changes in the color of the video frames (luma component).}
	\label{fig:system-overview}
\end{figure}
In recent years, the photoplethysmogram (PPG) has become particularly interesting in the context of mobile devices. PPG is a method to optically detect changes in blood flow volume and is often used in optical heart rate sensor of smartwatches. In the medical domain, PPG is used to monitor patients' heart rates and blood oxygen levels through pulse oximeters. Similarly to the electrocardiogram (ECG), PPG offers a range of distinctive biometric features derived from unique characteristics of an individual's cardiovascular system. Previous research in the field has relied on medical datasets or purpose-built PPG sensors. However, using these systems is not practical due to the specialized hardware required (for example, users are unlikely to clamp a pulse oximeter on their finger for authentication).

In this work, we are proposing the first system to collect a user's PPG through a mobile phone camera and authenticate them through the extracted signal.
During a measurement, the user places their finger on the phone's camera while the finger is illuminated by the phone camera flashlight (see Figure~\ref{fig:system-overview}).
Smartphone-based PPG authentication has two main advantages over established biometrics: Unlike fingerprints or facial images, PPG is hard to observe from a distance, making it less susceptible to presentation attacks. In addition, the hardware requirements are low and no specialized sensors are needed. This makes the technique even suitable for low-cost feature phones.
We validate our approach through a user study with 15 participants and capture their PPG over 6-11 sessions.
The multi-session nature of our experiment allows us to explore the time stability of the biometric, a component largely overlooked by previous work.
In order to facilitate future work in the field, we make our code and the dataset available online\footnote{https://github.com/ssloxford/seeing-red/}.

\section{Background and Related Work}

In this section, we will present the physiological foundations of PPG-based authentication, discuss different measurement methods and summarize related work.
\subsection{Physiological foundations}
A photoplethysmogram (PPG) is an optically obtained plethysmogram that can be used to detect blood volume changes in the microvascular bed of tissue~\cite{PPGDef}. 
Having first been proposed in 1938, PPG is a relatively simple technique that only relies on a light source to illuminate the skin tissue, and a photo detector to measure the small variations in light intensity associated with changes in perfusion in the catchment volume~\cite{maeda2008comparison}.
Within each heartbeat three distinct fiducial points can be identified that relate to different stages of the cardiac cycle (see Figure~\ref{fig:features1}).
Some biometric features used in this work (see Section~\ref{sec:feature_selection}) are directly based on these fiducial points and derive their distinctiveness from the uniqueness of the person's cardiovascular system.

The systolic peak is a result of the direct pressure wave traveling from the left ventricle to the periphery of the body~\cite{brumfield2005digital}. 
Its amplitude relates to stroke volume~\cite{murray1996peripheral} and is used to estimate continuous blood pressure~\cite{chua2010towards}.
The diastolic peak (or inflection) is a result of reflections of the pressure wave by arteries of the lower body~\cite{brumfield2005digital}.
The time difference between the systolic peak and diastolic peak has been previously used as a measure of large artery stiffness~\cite{millasseau2002determination}
The dicrotic notch is a small and brief increase in arterial blood pressure that appears when the aortic valve closes and is commonly used as an equivalent of end-systolic left ventricular pressure~\cite{dahlgren1991left}.
A more in-depth review of the PPG signal morphology and the medical reasons for the distinctiveness of individual components can be found in~\cite{elgendi2012analysis}.

\subsection{Measuring PPG}
\label{sec:devices}


\begin{figure}
     \centering
        \includegraphics[width=0.95\columnwidth]{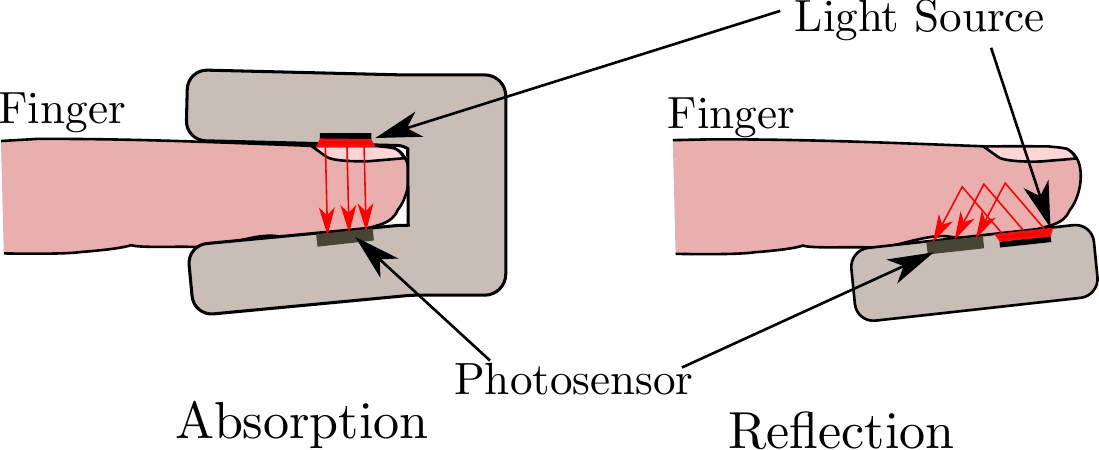}
        \caption{The two modes of PPG collection. The absorption mode is usually used by devices in a medical setting, such as pulse oximeters, whereas the reflection mode is used in smartwatches and other consumer systems. Our system operates in reflection mode.}
        \label{fig:ppg_hardware}
\end{figure}

In a clinical setting, PPG is typically measured with a pulse oximeter, usually clamped to a person's finger, and consists of a light source on the top and a photosensor on the bottom, as shown in Figure~\ref{fig:ppg_hardware}. The sensor then measures changes in light absorption, allowing it to distinguish between oxygenated and deoxygenated blood. 
A PPG sensor used in absorption mode must be located on the body at a site where transmitted light can be detected~\cite{Pereira2020}. 
Consequently, measurement sites are limited to the extremities of the body, such as the fingertips or earlobes.

In recent years, PPG sensors have become increasingly popular in wearable devices (smartwatches in particular) to allow continuous heart rate measurements without the need for a chest strap. 
PPG-enabled watches use a set of LEDs to illuminate the skin tissue and a photosensor to capture the resulting reflection, as seen in Figure \ref{fig:ppg_hardware}. 
The key difference to pulse oximeter measurements is that the light source and photosensor are adjacent. This is required as the absorption mode is not feasible on the wrist.

\subsection{PPG for Authentication }
While PPG was first proposed in 1938, it has only been used for authentication since 2003. Work in this area differs in terms of measurement device (particularly reflection vs absorption mode), sample size, feature types and consideration of feature stability over time. A summary of the field with regard to these properties can be found in Table~\ref{tab:rw}.

Most early work in the field uses relatively simple, low-dimensional feature sets~\cite{gu2003photoplethysmographic,gu2003novel}, focusing on the number of peaks and the slope of parts of the waveform. Bao et al. use heart rate variability (HRV) derived from PPG as a sole feature~\cite{bao2006physiological}.

More recent work has significantly extended the set of features to more comprehensively capture the distinctiveness of each constituent part of the PPG waveform and incorporates non-fiducial methods in order to avoid the difficulty of peak detection in noisy signals~\cite{karimian2017non}. The latter is particularly useful for the reflective measurement mode where the the diastolic peak and dicrotic notch are often hard to identify.


Unlike earlier work, which collected data from a single measurement session,  Kavsaoğlu et al. conducted three recording sessions with 30 participants~\cite{kavsaouglu2014novel}.
In order to compare intra-session and cross-session performance, they report individual results for each of the sessions and for a combined dataset consisting of both sessions.
The authors achieve identification rates of 90.44\% and 94.44\% for single-session tests and 87.22\% for the combined dataset.
However data from the sessions have not been separated in the evaluation, which positively biases the results compared to a real-world scenario~\cite{eberz2017evaluating}.


The most important difference in our work is the measurement device.
Previous work relies on either purpose-built hardware or medical-grade devices, whereas we use readily available consumer hardware. 
In addition, most research uses single-session data and performs training and testing within the same session. Due to the single-session nature of most experiments, it is possible that systems distinguish measurement sessions, rather than individuals. For example, PPG measurements are affected by skin temperature \cite{jeong2014effects} and sensor contact pressure and position~\cite{castaneda2018review}, both of which can easily introduce artifacts into the signal that would only be present in this recording. 

\begin{table}
\centering
\small
\begin{tabular}{ccccc}
\textit{Ref} & \textit{Measurement} & \textit{Subjects} & \textit{Dataset} & \textit{Sessions} \\ \toprule
   \cite{gu2003novel} & reflection & 17 & own & 1 \\
    \cite{gu2003photoplethysmographic} &  reflection & 17 & own & 1 \\
    \cite{bao2006physiological} & unknown & 12 & own & 1 \\ 
    \cite{yao2007pilot} & absorption & 3 & own & 1 \\
    \cite{spachos2011feasibility} & absorption & 14/15 & \cite{opensignal2} & 1 \\
    \cite{bonissi2013preliminary} & absorption & 44 & own & 1 \\ 
    \cite{kavsaouglu2014novel} & reflection & 30 & own & 3\\
    \cite{sarkar2016biometric} & absorption & 23 & \cite{koelstra2011deap} & 1 \\
    \cite{karimian2017non} & absorption & 42 & \cite{karlen2013multiparameter} & 1  \\
    \cite{karimian2017human} & absorption & 42 & \cite{karlen2013multiparameter} & 1 \\
    \cite{Zhao2020TrueHeart} & absorption & 20 & own & 1 \\
    \textbf{Our work} & reflection & 15 & own & 6-11 \\

    \bottomrule
\end{tabular}
\caption{Related work on PPG-based authentication\label{tab:rw}.}

\end{table}

\section{Experiment Design}
In this section we principally present our data collection method, which we subsequently use for developing our authentication systems.
\subsection{Collection Apparatus}
As previously mentioned, a key contribution of our work is the ability to use PPG for authentication without purpose-built hardware.
As such, our system makes use of a smartphone's camera and camera flash to act as the photosensor and light source respectively.
One challenge in this setup is presented by the camera light: commercial PPG sensors in reflection mode use green light as its wave length leads to ideal penetration depth~\cite{maeda2008comparison}.
Conversely, the phone camera uses white light, which leads to noisier measurements.
We collect the PPG data using a custom iOS application, running on an Apple iPhone X.

Users are instructed verbally to hold the phone in their dominant hand, and place their finger lightly over both the camera and camera flash.
They are instructed to sit down whilst performing the capture, and to remain as still as possible throughout; the application automatically stops recording if too much movement is detected i.e total acceleration exceeds 1.3g ($12.75 m/s^{2}$).

Additionally, we fix the camera ISO and exposure time to be the minimum possible, given the camera FPS and resolution settings.
Likewise, we set the white balance gain to be the maximum in the red channel, and minimum in the blue and green channels.
By fixing these settings we ensure consistency between capture sessions, reducing noise in the signal and increasing the information captured about the color changes.

A single capture session, as taken by the application, consists of a video recorded in 1280 x 720 resolution at 240 frames per second, for a duration of 30 seconds.
The video is resized to 360 x 240 and uploaded to a server for validation.
After each capture session, the researcher supervising the data collection takes the phone away from the user and returns it to them for any subsequent capture sessions, leading to users placing their finger on the camera differently each time.
A validation function analyses the red light channel of the video once it has been uploaded to the server, and rejects the video if there are sudden jumps or a lack of red light, as these are likely caused by the user lifting their finger from the camera. 

\subsection{Collection Procedure}\label{sec:data_collection}
We collect data for a total of 15 participants, performing between 6 and 11 capture sessions (measurements) per participant.
There were 13 male and 2 female participants.
Of the participants, 13 identify as white, one as Indian and one as South-East Asian.
Each participant is handed the device running the application and given the above instructions on how to use the application, as well as being given a visual prompt towards correct usage on the screen.
Captures for each participant were taken across multiple sittings, with no more than three capture sessions occurring in the same sitting.
Sittings were taken approximately a day apart, with a median of 32.1 hours between any pair of capture sessions.

\subsection{Research Ethics}
Before the data collection, we obtained informed consent from the participants.
Our data collection process and experiment took place with ethical approval from our institution: reference SSD/CUREC1A CS\_C1A\_19\_032.

\section{Method}
Our method consists of a pipeline of steps which transform the raw video captured by the camera into biometric samples which can be used for authentication.
We design the pipeline with the aim of running the system in real-time, meaning that all processing steps can be run as the video is being captured (i.e., we never use processing that requires looking at a whole video file, only at the video recorded up to that point).

\subsection{Signal Extraction}
To obtain the signal from the raw video, we compute the mean of the pixel-wise luma component from the pixels in each video frame, so that if $F$ is a video composed by a sequence of frames $\{f_1, ..., f_m\}$, then the signal originating from $F$ is:
\begin{equation}\label{eq:luma}
\begin{split}
S=\{Y(f_1), ..., Y(f_m)\}, \text{where} \qquad\qquad \\
Y(f) = \frac{1}{n} \sum_{i,j\in f} [ 0.299 f^{(r)}_{i,j} + 0.587 f^{(g)}_{i,j} + 0.114  f^{(b)}_{i,j}].
\end{split}
\end{equation}
In Equation~\ref{eq:luma},  $i$ and $j$ iterate over the pixels of the image and the superscripts $\_^{(r)}$ indicate the considered RGB channel of the frame, either red green or blue. 
The channel coefficients of Equation~\ref{eq:luma} are taken from the ITU-R BT.601 standard.\footnote{\url{https://en.wikipedia.org/wiki/Rec._601}}

\subsection{Signal Preprocessing}
\begin{figure}[t]
	\centering
	\includegraphics[width=\columnwidth]{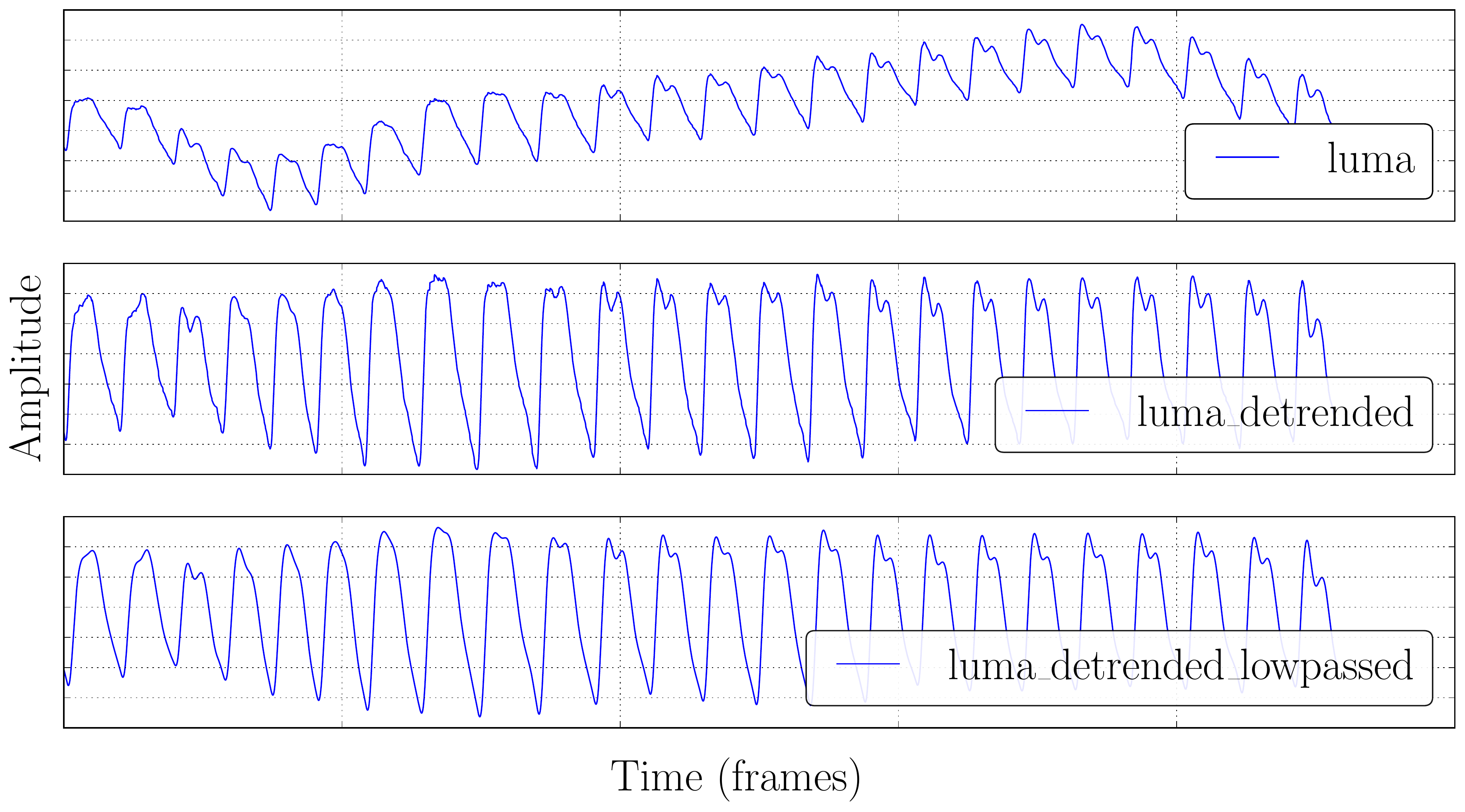}
	\caption{Signal over the course of the preprocessing steps. The first plot shows the luma component extracted by the video frames (Eq.\ref{eq:luma}). The second plot shows the signal after de-trending with rolling average and the third plot shows the signal after the low-pass filtering, which is used in the rest of the pipeline.}
	\label{fig:preprocessing}
\end{figure}
After the signal from a video is extracted, we apply the following preprocessing steps to remove noise.
First, in order to remove trends from the signal we compute and subtract a rolling average of the signal with the signal itself.
We use a window size of 1 second for the rolling average.
Then, we use a low pass filter to remove high frequency noise, with cutoff frequency at 4Hz, i.e., 240beats per minute.
See Figure~\ref{fig:preprocessing} for a visualization of the processed signals.

\subsection{Beat Separation}

As mentioned above, in our approach we want to perform single beat authentication, therefore we design a beat detection algorithm to separate individual beats in the signal.
We report in Algorithm~\ref{alg:beat_detection} the procedure to obtain individual beats from a preprocessed signal.
Alg.~\ref{alg:beat_detection} first finds the minimums of a smoothed version of the signal (moving average, Line 3), and then matches those minimums to the minimums in the original signal. 
The $\argrelmin{}$ instruction extracts all relative minimums indexes from the signal.
We found that this algorithm leads to a more reliable beat separation, as the signals often presents random noise which survives the preprocessing above.

\begin{algorithm}[t]
	\caption{\textbf{- Beat Separation.} Given a signal detects individual heartbeats boundaries.}\label{alg:beat_detection}
	\begin{algorithmic}[1]
		\State \textbf{Input:} preprocessed signal $S$, max\_bpm $U$, sampling rate $fps$, smoothing window size $ws$.
		\State $g$ = $60\cdot \frac{fps}{U}$ \Comment{minimum inter-beat gap}
		\State $A = \text{moving\_average}(S, ws)$
		\State $mins^{(S)} = \argrelmin{} S$
		\State $mins^{(A)} = \argrelmin{} A$
		\State $B = \{\}$ \Comment{Individual Beats}
		\For{$i \in mins^{(A)}$}
		    \State $S_i = \{s_{i-g}, ..., s_{i+g}\}$
		    \State $j = \argmin{{i-g}, ..., {i+g}} S_i$
		    \State $B = B + \{j\}$
		\EndFor
		\State \textbf{return} $B$
	\end{algorithmic}
\end{algorithm}

\subsection{Fiducial Points Detection}
After individual beats have been extracted, we detect fiducial points in the signal, which are used to extract features.
We focus on three points in particular: (i) systolic peak, (ii) dicrotic notch and (iii)  diastolic peak.
In low-noise signals, these points are easily found by finding maximums and minimums in the signal and its first derivative, see Figure~\ref{fig:features1}.
However we found that this procedure needs to account for noisy signals, so we craft a more robust algorithm to detect them which falls back to best guess points.

\subsection{Beat Signal Quality}\label{sec:beat_signal_quality}
Oftentimes, slight hand or finger movements may cause noise in the signal which bypasses our filtering pipeline.
We therefore design a set of quality criteria for individual beats with the goal of excluding noisy beats from further processing.
These criteria are the following:
\begin{itemize}
    \item \textbf{max bpm}: all beats which correspond to $>120$ bpm, as these are caused by noise which bypasses the separation of Alg.~\ref{alg:beat_detection};
    \item \textbf{number of peaks}: all beats with more than three distinct maximums;
    \item \textbf{distance from reference}: all beats that are too different from a reference expected beat wave. We compute this difference using dynamic time warping~\cite{sakoe1978dynamic} between the current beat and the reference and we set a threshold of 2.0, i.e., beats with distance higher than 2.0 are discarded. The reference wave is the average beat wave in our dataset.
\end{itemize}
All beats that match at least one of these conditions are flagged as failed to acquire (FTA).
Using these quality filters leads to better authentication accuracy at the cost of slightly longer authentication time. In the analysis, we report the results including and excluding these beats.

\subsection{Feature Extraction}\label{sec:feature_extraction}

\begin{figure}[t]
	\centering
	\includegraphics[width=\columnwidth]{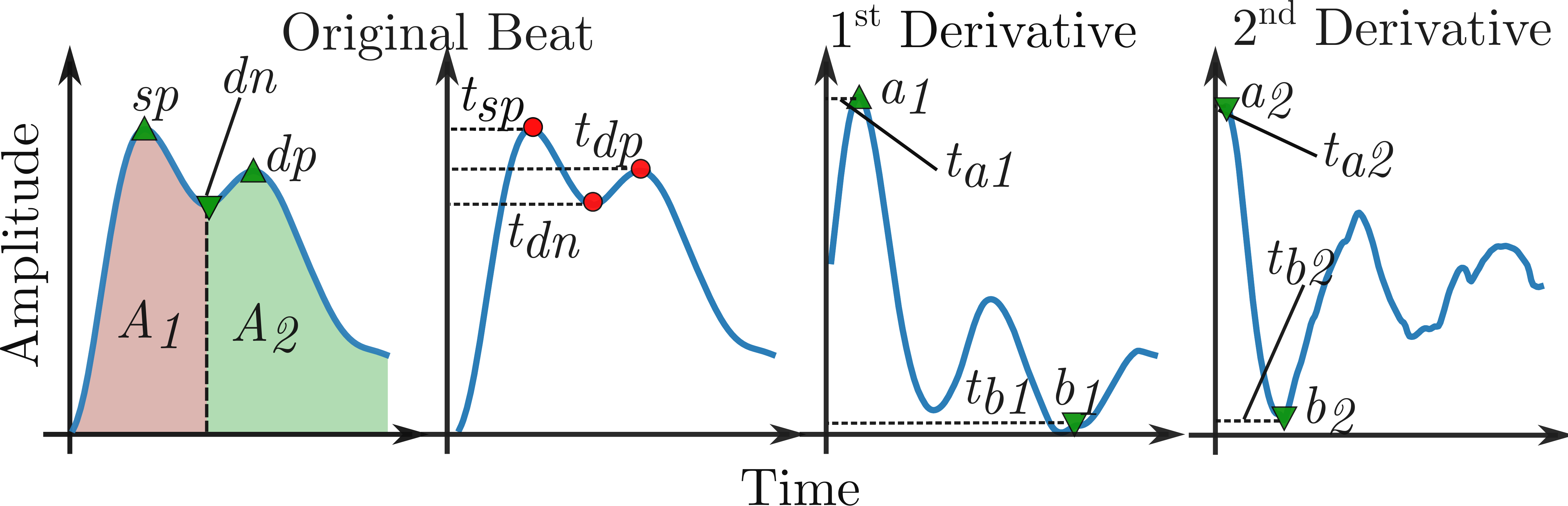}
	\caption{Visualization of fiducial point based features for a single beat. Time span features are based on time from start of an individual heartbeat, and are normalized to account for different bpm.}
	\label{fig:features1}
\end{figure}

We consider four groups of features: statistical, curve widths, frequency domain and fiducial points based.
To make features independent of a user's bpm at the time of capture, after the physiological features are computed, we re-sample beats onto a fixed sampling rate of 1,000 Hz and we normalize them so that the amplitude values lie in [0, 1].
Features are computed on a beat-basis, so that each beat leads to a sample.

\myparagraph{Statistical}
These are straightforward statistical features that are based on the physiological traits of the users and for a single beat include: (i) maximum value, (ii) minimum value, (iii) the difference max-min and (iv) the length of the beat.
Given that our camera settings are fixed, features (i)-(iii) are a result of the absolute values found in the luma component of the video frames, which are related to skin tone and amount of blood flowing through the finger.
Feature (iv) is just related to the individual's heart rate at rest, which has some variability, but we found it to be a distinctive features in our case.

\myparagraph{Curve Widths}
These features are the widths of the curve defined by the beat at a set of pre-defined heights.
For a certain height $h \in [0, 1]$, where $1$ is the maximum height and $0$ the minimum, we consider the curve that lies above the threshold $h$ and compute the distance between the extremes of this curve, see Figure~\ref{fig:features1}.
We choose 18 different heights evenly spaced in [0.05, 0.95] and compute one width feature for each of them.

\myparagraph{Frequency Domain}
For each beat we compute the discrete Fourier transform and use it as features.
Section~\ref{sec:feature_selection} discusses how we reduce the dimensionality of these features with feature selection.

\myparagraph{Fiducial Points}
We add a set of features based on the location of the fiducial points discussed above based on related work, see Figure~\ref{fig:features1} for examples of the features used.

\subsection{Feature Selection}\label{sec:feature_selection}
After feature extraction, each beat is represented by a total of 541 features, 4 physiological, 18 curve widths, 500 frequency domain and 19 from the fiducial points.
To reduce dimensionality and remove redundant features, we apply multi step feature selection.

\myparagraph{Step 1} 
We use principal component analysis (PCA)~\cite{tipping1999probabilistic} to reduce the number of features in the frequency domain and the curve width groups.
For the frequency features, we fit a PCA with 100 components and retain only the first $n$ components so that these $n$ components describe 99\% of the space variance.
For the curve width features, we do the same but fit a PCA model with 15 components.
We find that roughly 5 components are necessary for the frequency group (depending on the dataset) while 9 are sufficient for the width group.

\myparagraph{Step 2} 
In this step, we combine two different techniques for feature selection on the remaining features.
At first, we compute the pairwise correlation coefficient between features and for a pair of feature distributions $(f_1, f_2)$ we drop one feature from the dataset at random if the correlation coefficient between them is $r_{f_1, f_2} >.95$.
Then, to avoid the effect of outliers, we remove all feature values which lie outside the $1^{\text{st}}$ and $99^{\text{th}}$ percentile of the feature distribution.
Then we use minimum redundancy maximum relevance (mRMR) feature selection~\cite{peng2005feature} to select the top 60\% performing features (differently from the original paper, we choose the Mutual Information Quotient criteria in the algorithm).
Afterwards, we select the top 60\% features ranked on their relative mutual information (RMI). 
We use the non-parametric approach to estimate RMI introduced in~\cite{Ross2014} to avoid imprecisions arising from modeling continuous distributions with discrete bins.
Finally, we choose all the features 
that pass both selections (RMI and mRMR) to create our final feature set.
Section~\ref{sec:results} presents and discusses the selected features set.

\section{Results}
\label{sec:results}
In this section we present the results for a set of authentication use-cases.

\myparagraph{Preliminaries}
In each use-case, we mainly monitor the equal error rate (EER) of the authentication, that is the point at which False Reject Rate (FRR) and False Accept Rate (FAR) are equal.
In each experiment, rather than using the classification output from the classifier as the decision (accept/reject), we use the distance from the decision boundary (or prediction probability) to make the classification decision, i.e., we list all distances coming from the positive (the user) and negative (other users) samples, and find the EER by changing the threshold.

\myparagraph{Aggregation} 
We report results for aggregating the decision on multiple samples, i.e., we choose an aggregation window size $n$, collect $n$ samples and compute the decision based on the aggregated samples.
The aggregation function we use is the mean.\footnote{We also tried median but it performed worse on average.}
To gather more data points, whenever we aggregate samples together, we randomly draw sample sequences of length $n$ for 100 times from the positive (genuine) class.
For a certain user, to construct the negative (impostor) class, we randomly draw sample sequences of length $n$ for 10 times for each other user.
This way, each EER value for a user is computed using $100+10\times(|U|-1)=240$ authentication attempts.

\begin{table}[tp]
    \centering
    \small
    \begin{tabular}{c|c|c}
        \textit{Feature} & \textit{mRMR} (MIQ) & \textit{RMI} \\ \toprule 

        $A_{2}/A_{1}$ & 1.29 & 0.14 \\ 
        $t_{b_1}$ & 1.11 & 0.13 \\ 
        $b_{}/a_{2}$ & 1.11 & 0.10 \\ 
        $b_2$ & 1.02 & 0.13 \\ 
        $max$ & 1.02 & 0.17 \\ 
        $min$ & 0.88 & 0.23 \\ 
        $a_2$ & 0.86 & 0.12 \\ 
        $t_{b_2}$ & 0.82 & 0.12 \\ 
        $t_{sp}$ & 0.77 & 0.12 \\ 
        $\text{fft}_{PCA_0}$ & 0.75 & 0.09 \\ 
        length & 0.72 & 0.12 \\ 
        $\text{fft}_{PCA_1}$ & 0.70 & 0.08 \\ 
        $t_{dn}$ & 0.69 & 0.15 \\ 
        $t_{a_2}$ & 0.67 & 0.08 \\ 
        $\text{fft}_{PCA_5}$ & 0.66 & 0.08 \\ 
        $b_1$ & 0.64 & 0.12 \\ 
        $t_{a_1}$ & 0.64 & 0.10 \\ 
        $A_1$ & 0.61 & 0.11 \\ 
\bottomrule
    \end{tabular}
    \caption{Features retained and their importances after the feature selection step, for both ALL and post-FTA datasets. The importances reported are computed in the ALL dataset.}
    \label{tab:table_features}
\end{table}
At the end of our data collection, we obtain a dataset of a total of 3,836 samples (one sample equals one beat) with at least 178 samples per user.
Of the samples, 3,529 (92\%) pass the beat signal quality checks  discussed in Section~\ref{sec:beat_signal_quality}.
Throughout the evaluation, we refer to these two separate datasets as \textbf{ALL} and \textbf{Post-FTA} and we report results for both.
After feature selection, we retained the top 18 features for both ALL and Post-FTA, see Table~\ref{tab:table_features} for details.


\subsection{Multi-class Case}
\begin{figure}[t]
	\centering
	\includegraphics[width=.95\columnwidth]{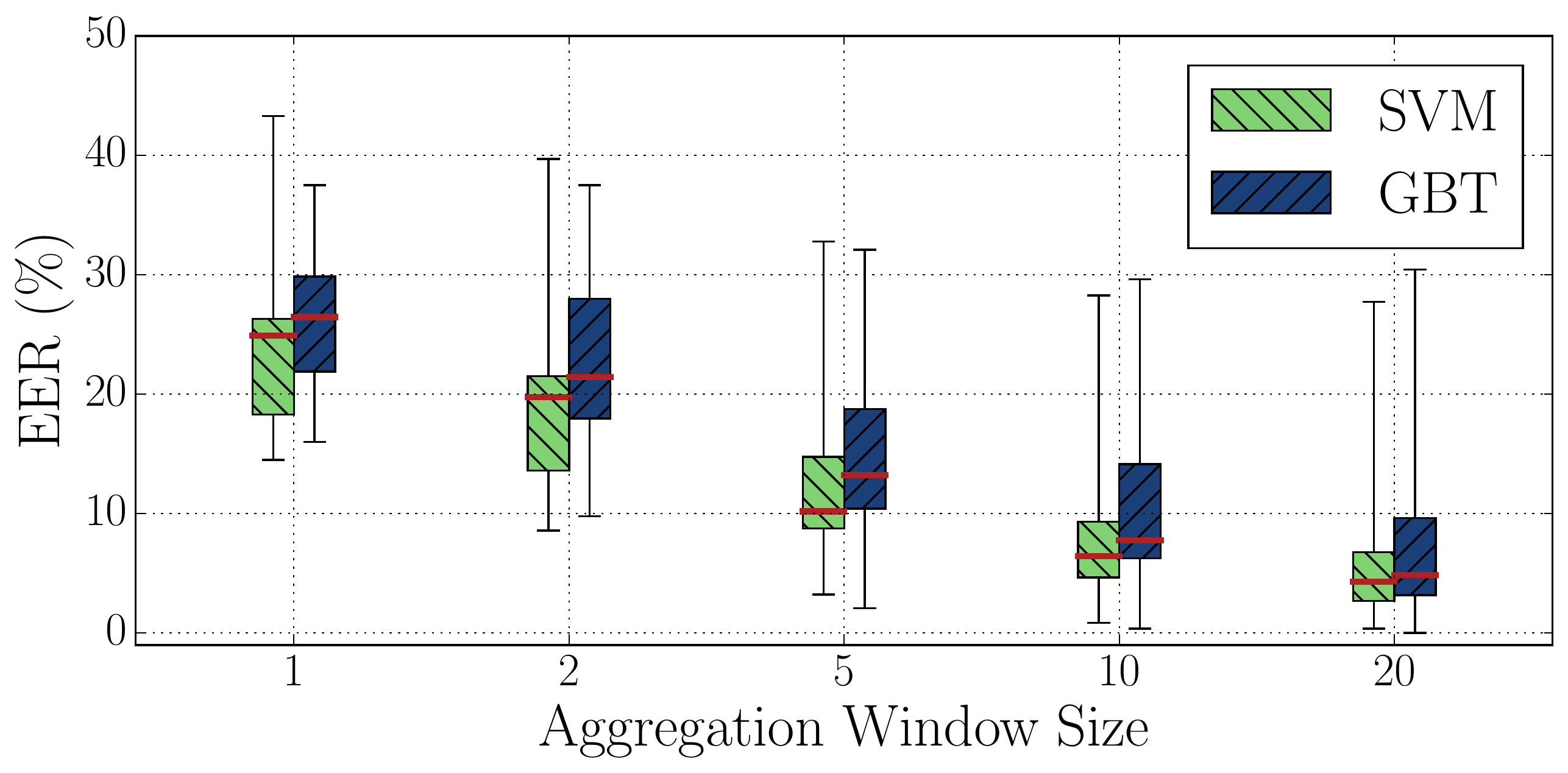}
	\caption{EERs in the multi-class case, for various classifiers and aggregation window sizes. Values are averaged over the two folds before plotting. Whiskers show the range of the data, horizontal lines in boxes show the median.}
	\label{fig:multiclass}
\end{figure}

\myparagraph{Setup}
We randomly split the data into two stratified folds and train two different classifiers, support vector machine (SVM) with a radial basis function kernel and gradient boosted trees (GBT).
For SVM we use a standard scaler to normalize the data before feeding it into the classifier, and only use the training part to fit the scaler.

\myparagraph{Results}
We report in Figure~\ref{fig:multiclass} the obtained EERs.
We found that single sample EER lies around 15\%, but decreases quickly to less than 5\% when using an aggregation window size $>10$.
Out of the tested models we found that SVM performs the best, with an EER of \textless1\% at aggregation window size of 20.

\subsection{One-class Case}\label{sec:one_class_case}
\begin{figure*}[t]
	\centering
	\includegraphics[width=\textwidth]{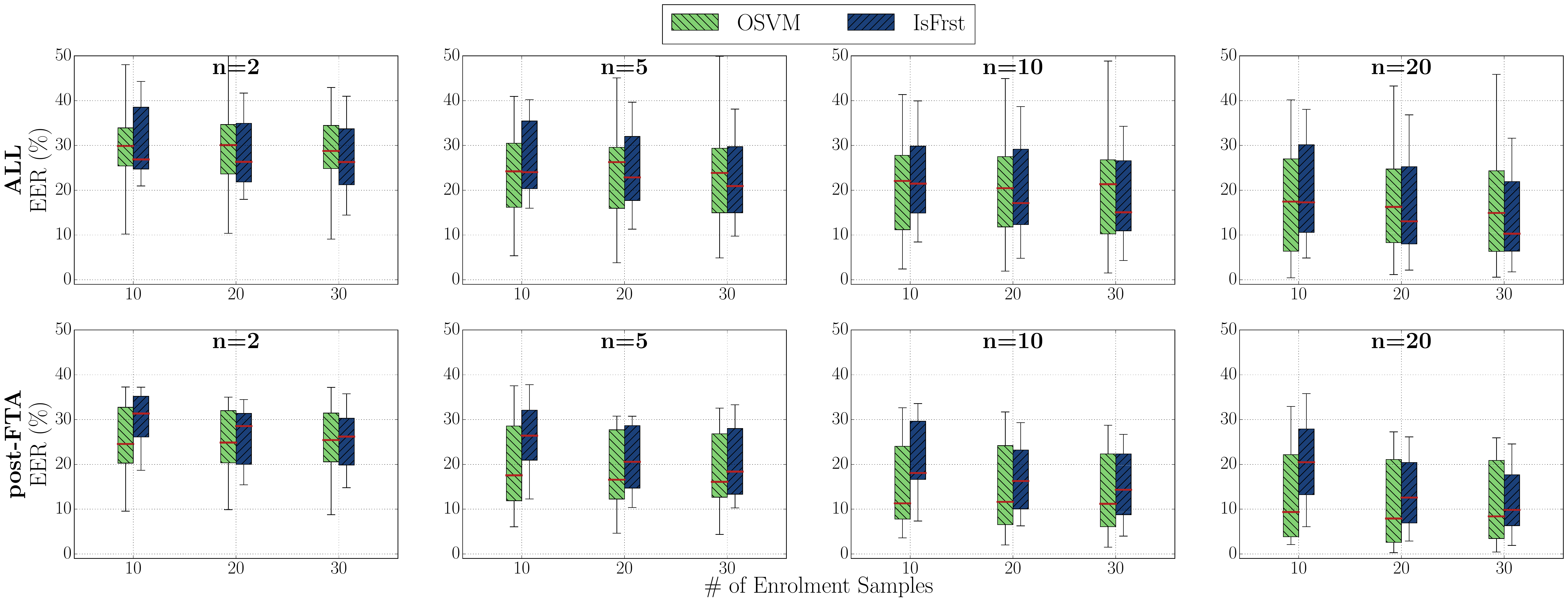}
	\caption{EERs in the one-class case, for various classifier, aggregation window sizes and no. of enrollment samples. Values are averaged over the number of random enrollments first and then across users. The first row of plots reports the results for the ALL dataset, while the second row reports the post-FTA dataset. Each plot reports the aggregation window size $n$ used at the top, i.e., either 2, 5, 10 or 20. Whiskers show the 95\% confidence intervals.}
	\label{fig:exp3eer}
\end{figure*}
\myparagraph{Setup}
As a PPG-based authentication system leveraging the smartphone camera would reside on the user's device, the system would likely only have access to the legitimate user's data (provided at enrolment).
We therefore consider the case where the training algorithms do not have access to  any negative class data.
We use two algorithms which can be fit only to a single class: Isolation Forest (IsFrst)~\cite{liu2008isolation} and a one class SVM (OSVM).
The training data is therefore only provided at enrolment, when the user first uses the system, and as such we vary the number of samples provided at enrolment.
This helps us provide insights on how many samples are necessary to capture sufficient intra-user variation.
The rest of the experiment follows the same method described at the beginning of Section~\ref{sec:results}, with the difference that we repeat the enrolment 10 times to ensure that the enrolment samples do not bias the results (selecting particularly good/bad samples for enrolment will greatly influence the results).
To compute the EER, we use the distance from the decision boundary for OSVM and the anomaly score for IsFrst.
The results are averaged over the number of repeated enrolments first and then over the individual users. 

\myparagraph{Results}
We report the results of the analysis in Figure~\ref{fig:exp3eer}.
The figure shows the distributions of EERs across: (i) ALL or Post-FTA dataset, (ii) number of enrolment samples, (iii) aggregation window size.
The number of enrolment samples represents a small fraction of the available user data (for each user we have at least 178 samples, and we use as few as 10 for training).
We found that OSVM performs slightly better on average but not in a consistent manner.
Using the Post-FTA dataset leads to better recognition results (as low as 4\% at 20 aggregated samples) and smaller variations in the EER distributions, while the average EER for the ALL dataset is above 10\% for most configurations with the exception of using IsFrst and aggregating 20 samples.

\subsection{One-class Cross-Session}
\myparagraph{Setup}
The evaluation methodology is identical to the one described in Section~\ref{sec:one_class_case}, with the difference that now we select enrolment data by choosing a number of sessions (we have at least 6 sessions per user, with a mean of 9.3 sessions and median of 10 sessions) and test on the remaining sessions. A key difference to the work by Kavsaouglu et al.~\cite{kavsaouglu2014novel} is that we draw training and testing samples from separate sessions. This makes our analysis more realistic, as our results are not positively biased by features that are not stable between sessions.

\myparagraph{Results}
\begin{figure*}[t]
	\centering
	\includegraphics[width=\textwidth]{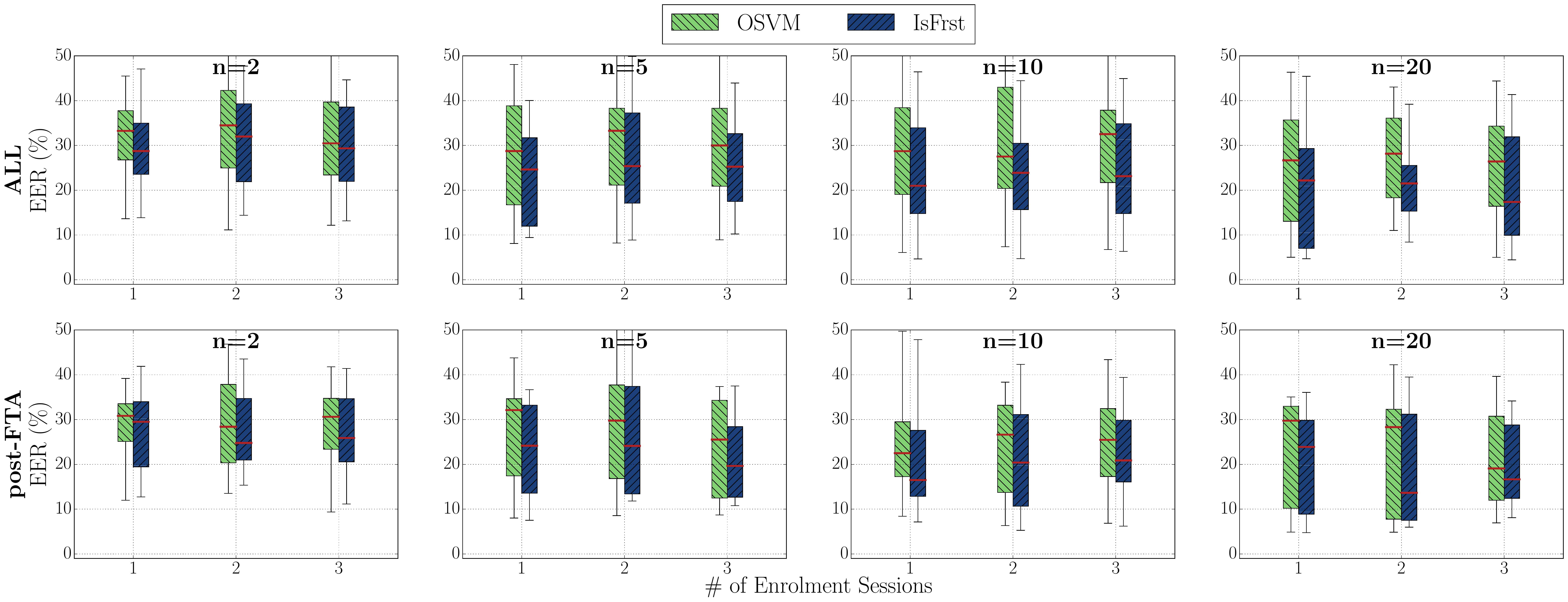}
	\caption{EERs in the one-class cross-session case, for various classifier, aggregation window sizes and no. of enrolment sessions. Values are averaged over the number of random enrolments first and then across users. The first row of plots reports the results for the ALL dataset, while the second row reports the post-FTA dataset. Each plot reports the aggregation window size $n$ used at the top, i.e., either 2, 5, 10 or 20. Whiskers show the 95\% confidence intervals.}
	\label{fig:exp4eer}
\end{figure*}
We report the results of the analysis in Figure~\ref{fig:exp4eer}.
As in the previous section, the figure shows the distributions of EERs across: (i) ALL or Post-FTA dataset, (ii) number of enrolment sessions, (iii) aggregation window size.
As mentioned in Section~\ref{sec:data_collection}, we have 6-11 measurement sessions per user, the number of enrolment sessions (which varies between 1-3) represents between 10\%-50\% of available data per user.
We found that the EERs are much higher in this case, with most scenarios scoring over 20\% EER on average and high variance in the results.
This suggests that each PPG measurement is not only unique to the user, but also to the way the measurement is being taken, leading to most of our features being inconsistent over separate sessions.

\subsection{EER User Distribution}
\begin{figure*}[t]
	\centering
	\includegraphics[width=\textwidth]{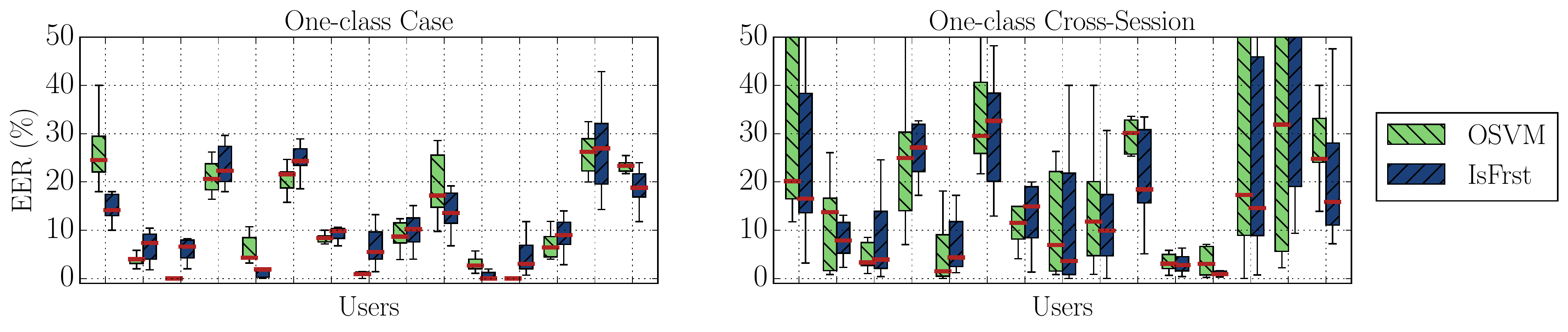}
	\caption{Distribution of EERs across users in our dataset, for the one-class and cross-session case, the Post-FTA dataset and an aggregation window size of 20. Whiskers show the 95\% confidence intervals.}
	\label{fig:eer_distrib}
\end{figure*}
We investigate how EER distributes over different users, showing the results in Figure~\ref{fig:eer_distrib}.
The figure shows how the distribution is uneven across users, with the worst performing user contributing the most to the overall EER. 
User 6 has over 20\% EER in the One-class case and above 30\% in the cross session case, while the authentication is relatively stable for a subset of users even in the cross-session scenario (e.g., User 3, 5, 11, 12 have below 5\% EER).
We found that such results are highly susceptible to the quality of samples used for enrolment, suggesting that stricter quality metrics in this phase could further improve robustness.
Equally this effect could be caused due to specific, unknown characteristics of given individuals, as has been shown to be the case in biometric systems previously~\cite{Doddington1998SHEEPGL}.


\section{Conclusion}

In this paper, we proposed a system that enables PPG-based authentication by using a smartphone camera.
The PPG signal is collected by recording a video from the camera as the user is resting their finger on top of the lens. 
We extract the signal based on subtle changes in the video that are due to changes in the light absorption properties of the skin as the blood flows through the finger.
Using an iPhone X, we collect a dataset of PPG measurements from a set of 15 users, over the course of 6-11 capture sessions per user.
We design a pipeline for the data analysis and conduct a set of experiments to evaluate the recognition performance based on a set of features extracted from individual heartbeats.
We found that when aggregating sufficient samples for a decision we can achieve equal error rates as low as 8\%, but that the performance greatly decreases in the cross-session scenario, with EERs around 20\% on average across all users.

\myparagraph{Future Work}
We observed that a set of factors greatly affect the fidelity of the signal.
At first, we found that the warmth of the fingertip changes the amount of blood flow, leading to slightly different measurements.
Secondly, breathing has an effect on the signal: heartbeats are stronger when inhaling compared to exhaling~\cite{barschdorff1994respiratory}.
Additionally, slight hand and/or finger movements generate noise that can be (in part) corrected by using accelerometer and gyroscope measurements.
We also plan to better understand the device capabilities needed for this technique to be used successfully.
Firstly, evaluating the hardware layout impact on the signal quality, in particular distance between light source and camera.
Secondly, using lower frame rates and lower resolution  recordings, as using more coarse-grained measurements will enable the use of PPG authentication on lower-end devices.
Finally, a larger dataset, both in terms of users and sessions will help us draw more insights on the time stability of the PPG biometric trait.

\section*{Acknowledgements}\noindent
This work was generously supported by a grant from Mastercard and by the Engineering and Physical Sciences Research Council [grant numbers EP/N509711/1, EP/P00881X/1].

\balance

{\small
\bibliographystyle{ieee_fullname}
\bibliography{references}

}

\end{document}